\documentclass{article}

\usepackage{arxiv}

\usepackage[utf8]{inputenc} 
\usepackage[T1]{fontenc}    
\usepackage{hyperref}       
\usepackage{url}            
\usepackage{booktabs}       
\usepackage{amsfonts}       
\usepackage{amsmath}
\usepackage{nicefrac}       
\usepackage{microtype}      
\usepackage{graphicx}
\usepackage{doi}
\usepackage{todonotes}
\usepackage{algorithm}
\usepackage{algpseudocode}

\newcommand{\romnum}[1]{\uppercase\expandafter{\romannumeral #1\relax}}

\title{Positivity Validation Detection and Explainability via Zero Fraction Multi-Hypothesis Testing and Asymmetrically Pruned Decision Trees}

\author{ Guy Wolf\\
	Vianai Systems\\
	\texttt{guy.wolf@vian.ai} \\
	\And
	Gil Shabat \\
	Vianai Systems\\
    \texttt{gil@vian.ai} \\
    \And
	Hanan Shteingart \\
	Vianai Systems\\
    \texttt{hanan@vian.ai}
    }

\renewcommand{\shorttitle}{Positivity Validation and Explainability}

\begin{document}
\maketitle

\begin{abstract}
	Positivity is one of the three conditions for causal inference from observational data. The standard way to validate positivity is to analyze the distribution of propensity. However, to democratize the ability to do causal inference by non-experts, it is required to design an algorithm to (i) test positivity and (ii) explain where in the covariate space positivity is lacking. The latter could be used to either suggest the limitation of further causal analysis and/or encourage experimentation where positivity is violated. The contribution of this paper is first present the problem of automatic positivity analysis and secondly to propose an algorithm based on a two steps process. The first step, models the propensity condition on the covariates and then analyze the latter distribution using multiple hypothesis testing to create positivity violation labels. The second step uses asymmetrically pruned decision trees for explainability. The latter is further converted into readable text a non-expert can understand. We demonstrate our method on a proprietary data-set of a large software enterprise. 
\end{abstract}

\vspace*{\fill} 
\begin{quote} 
\centering 
\it ``Positive anything is better than negative nothing.''  / Elbert Hubbard
\end{quote}
\vspace*{\fill}


\section{Introduction}
\label{sec:introduction}
Causal inference from observational data revolves around the hope for identification, meaning transforming the causal problem into a statistical one, or in other words viewed as a conditionally randomized experiment \cite{hernan2020causal}. 
In theory, there are three conditions that must hold so that an observational study can be conceptualized as a conditionally randomized experiment: (1) The values of treatment under comparison correspond to well-defined interventions, formally known as \textit{Stable Unit Treatment Value Assumption} (SUTVA), (2) The conditional probability of receiving every value of treatment, though not decided by the investigators, depends only on the measured covariates, formally known as \textit{Ignorability} or \textit{No Hidden Confounder} assumption), (3) The conditional probability of receiving every value of treatment is greater than zero, formally known as \textit {Positivity}. 

The positivity condition, which is the focus of this paper, can be mathematically expressed as $Pr(T=t|X)>0 \quad \forall{t}$, where $T$ is the treatment random variable, $t$ is the treatment and $X$ the observed covariates. For the rest of the paper, we assume $T \in \{0,1\}$ for simplicity, but the suggested methods can be generalized to multiple treatment.  As explained above, meaningful causal analysis can not be performed on areas of the data that lack positivity. This violation problem becomes more common in high dimensional covariate space \cite{d2021overlap}. To validate positivity, one would theoretically test whether there is a volume in space $X$, where at least one treatment type is missing. Thus, the task of finding a lack of covariate overlap can be translated into the task of checking whether the distribution of covariates is similar between the two treatment groups, formally $P(X|T=0)>0$ when $P(X|T=1)>0$ and vice versa. Yet, most statistical similarity tests do not scale naturally for multivariate distributions \cite{karavani2019discriminative}. Moreover, multivariate density functions cannot be assessed easily to help observe how distributions of the two groups differ. A low dimension alternative approach to mitigate this ``curse of dimensionality'' is to use propensity scores as a balancing score $\pi(X)=Pr(T=1|X)$. From a theoretical perspective \cite{rosenbaum1983central}, it is possible to compare the distribution of propensity scores, rather than covariates, between groups and alert for violations once the mass of the two distributions do not overlap, especially their tails. Such propensity scores can be estimated using a traditional machine learning classifier model, also known as a Propensity Model. In other words, violations of the positivity assumption are indicative of overlap problems in the distributions of the propensity scores, and consequently, of the observed confounders, among the treatment groups. Inference for subjects that lie off of the common support of the propensity scores is difficult and often requires extrapolation methods. A common method of addressing this overlap concern and identifying “non-positivity” subjects is to truncate the distribution of propensity scores so that only those subjects whose propensity scores lie on the common support are included in the analysis \cite{kang2016practice}.

Another way to look at the importance of overlap is as follows. Propensity score (PS) weighting methods are often used in non-randomized studies to
adjust for confounding and assess treatment effects. The most popular among them, the inverse probability weighting (IPW), assigns weights that are proportional to the inverse of the conditional probability of a specific treatment assignment, given observed covariates. A key requirement for IPW estimation is the positivity assumption, i.e., the PS must be bounded away from 0 and 1. In practice, violations of the positivity assumption often manifest by the presence of limited overlap in the PS distributions between treatment groups. When these practical violations occur, a small number of
highly influential IPW weights may lead to unstable IPW estimators, with biased estimates and large variances \cite{zhou2020}. In practice, two important diagnostic evaluations are conducted when using IPW. One, to evaluate the covariate balance (before and after weighting) and ensure that weighting leads to comparable treatment groups, with respect to the measured covariates. The other, to assess the positivity assumption by looking at the overlap of the PS distributions between the treatment groups and their common support, as done here.

Albeit important, the detection of an positivity violation by itself is limited as it does not allow an analyst to understand the limitation of the analysis nor to encourage further experimentation where positivity is violated. Therefore, an explanation of the properties of $X$ within the area is also required. A toolkit for the evaluation of propensity models, which attempts to emulate an RCT, by modifying the training sets of models has been previously suggested \cite{shimoni2019evaluation}. As a part of the toolkit, methods for the evaluation of the propensity of datasets are proposed. Their methods, however, only concern themselves with how close the dataset is to an RCT and do not provide a way to easily explain areas of nonpositivity within the data, which propensity methods are not able to correctly fix.

Positivity analysis, as described above, attracts researchers for being grounded in theory \cite{karavani2019discriminative}. However, it has been suggested there are two main drawbacks which we decline \cite{karavani2019discriminative}. Firstly, the authors claim that propensity scores can be viewed as a dimensionality reduction from the covariate space onto a single number, and almost always bear some information loss, yet the author refers to a paper on Principle Component Analysis (PCA) which is an unrelated unsupervised method. Moreover, the propensity score is known to be a balancing score, meaning it can produce an unbiased estimate for treatment effects even in the absence of randomization. Secondly, the author claim that once a violation is detected, it is usually hard to characterize the covariate subspace causing it, as it depends on the interpretability of the model used to obtain those propensity scores. By contrast, in this paper we show that explainability is possible for the detected violations and it is not explicitly related to the propensity model. 

\section{Suggested Method}
\label{sec:ourmethod}
We suggest a method for the detection of positivity violation and explainability within a given dataset.  The method consists of two steps and several sub-steps, which will be detailed in the following subsections. The two steps are Violation detection (\ref{subsec:detection}) and Explainability (\ref{subsec:explanation}). Violation detection, see Algorithm \ref{alg:detect}, includes propensity modeling (\ref{subsec:prop_model}), distribution estimation (\ref{subsec:hist}) and statistical violation detection (\ref{subsec:violation}), while Explainability includes violation modeling (\ref{subsec:model}), pruning (\ref{subsec:prune}) and text generation (\ref{subsec:exp_text}).

\subsection{Detectability}
\label{subsec:detection}

\begin{algorithm}
\caption{Violation Detection}
\begin{algorithmic}
\Require $X_{n \times d}$ where $n$ is number of samples and $d$ the feature dimension and a vector of treatment $T_{n
\times 1}$
\Ensure p-values for positivity violations, corrected for multiple hypothesis testing

\State $\pi_{n\times 1} \gets \texttt{fit\_predict}(X,T)$
\Comment Fit and predict a propensity, see \ref{subsec:prop_model}

\State $ h_{H \times 1}^{T=t} \gets \texttt{density\_est}(\pi, T)$
\Comment Estimate propensity density, see \ref{subsec:hist}

\State $v_{V \times 1} \gets \texttt{violation\_index}(h^{T=0}, h^{T=1})$
\Comment Detect violation bins, see \ref{subsec:violation}

\State $p_{V \times 1}(i) \gets \texttt{stat\_test}(h_i^{T=0}, n_{T=0}, h_i^{T=1}, n_{T=1}) \quad \forall{i}\in v$
\Comment Statistical test per violation, see \ref{subsec:violation}

\State $p'_{V \times 1} \gets \texttt{multi\_test}(p)$
\Comment correct for multiple tests, see \ref{subsec:violation}

\State $V_{H \times 1}(i) = \begin{cases}
    1, & \text{if } i \in v \quad \& \quad p'(i) < \alpha \\
    0,              & \text{otherwise}
\end{cases} \quad \forall{i \in \{1,H}\}$
\Comment significance as labels for explainability step, see \ref{subsec:explanation}
\end{algorithmic}
\label{alg:detect}
\end{algorithm}

\subsubsection{Propensity Modeling}
\label{subsec:prop_model}
We first model the propensity of the dataset using a classifier trained to differentiate between the treated and control sets using the covariates of the data as the features. Propensity scores are then extractable by running the classifier over the entire set to predict $Pr(T=1|X)$. In general the propensity model accuracy does not have to be perfect \cite{alam2019should}. Specifically, in order for the proposed method to work, the classifier does not need to be perfect since statistical tests that are applied later allow for some error in this stage.
The following two methods can be used to test the performance of the classifier in order to confirm that it meets basic accuracy standards:

\begin{enumerate}
    \item Using cross validation, the classifier can be checked for its predictability using a standard evaluation function such as cross entropy loss, area under the curve (AUC) \cite{zhang2007use}, etc.
    \item If the datapoints displaying positivity violations are known in advance, the classifier can be checked for accuracy on these points alone using an evaluation function.
\end{enumerate}

While the former method is a general method capable of being used on any dataset irrelevant to prior information, it also carries the risk of ignoring problems inherent to the non-positive datapoints. As such, a classifier should be tested on a dataset with a known positivity violation first before using an unknown dataset.

\subsubsection{Distribution Estimation}
\label{subsec:hist}

The propensity classifier is applied to all points on both sets, which yields a discrete set of propensity score values. In order to detect positivity, however, a discrete set is not enough - the distribution $h^{T=t}(\pi) = Pr(\pi|T=t)$ needs to be estimated (see Introduction). In order to estimate the distribution of the propensity scores, the area in which they might exist $[0,1]$ is split into bins of equal size. The data is then sorted into the bins, with the amount of data in the bins being used as an estimation for the distribution. This method has been found to practically work with $100$ bins each consisting of one percent of the propensity area. The method can be easily extended to non-equal bins. Formally, $h_i^{T=t}$ is the number of histogram counts in bin $i$ ($1\leq i \leq H$) and $H$ being the number of histogram bins for the $T=t$ group. 

\subsubsection{Violation Detection}
\label{subsec:violation}

After the treatment is modeled via a propensity model and the resultant propensity score distribution is estimated using a histogram for both the control and treated sets (see Figure \ref{fig:fdr} orange and blue bars, respectively), we need to find the areas in which violation is suspected. 
This is done explicitly by finding bins in which one of the distribution is zero and one is not, or formally $v = \left\{i \ni (h_i^{T=0}>0)\oplus(h_i^{T=1}>0)=1\right\}$ where $v$ is violated bins set 
and $\oplus$ is the \texttt{xor} (Exclusive-OR) logic operator. The latter is depicted as black ticks in Figure \ref{fig:fdr}.

A two-sample proportion hypothesis test \cite{fleiss2013statistical} is then conducted on each suspected violation in $v$ using the tuple $\left ( k_0=h_i^{T=0}, n_0=n^{T=0}, k_1=h_i^{T=1}, n_1=n^{T=1}\right ) $ where $k_{t}$ and $n_{t}=\sum_i{h_i^{t}}$ corresponds to the $i$-th bin count and total sample size of population $T=t$, respectively. Any other contingency table statistical test such as Fisher's exact test \cite{fisher1922interpretation} would also be applicable. The justification for the above test can be explained by considering the non-zero counts as a false positive rate $p_+=\frac{k_t^{\textit{nz}}}{n_t^{\textit{nz}}}$ where $t^{\textit{nz}}$ is the non zero count group \cite{bradley2013testing}.  Under the null hypothesis that the set is empty, that is, the probability of obtaining an element is $p = 0$, an imperfect classification process is the only way to obtain a positive count. Thus, the number $k_t^\textit{z}=0$ is a binomial random variable with parameters $n_t^\textit{z}$ and $p_+$ where $t^\textit{z}$ is the zero count group.

The test $p$-values per each bin in $v$ can then be compared to a significant level, e.g.  $\alpha=1\%$, as depicted in Figure \ref{fig:fdr} as green $X$. However, such a naive statistical test is invalid because one needs to take into account the fact that there are multiple hypothesis testings. Thus, we use the false discovery rate (FDR) procedure, a method of conceptualizing the rate of type \romnum{1} errors in null hypothesis testing when conducting multiple comparisons \cite{benjamini1995controlling}. The rejected hypothesis of equal distribution after the FDR procedure, as depicted in Figure \ref{fig:fdr} by red circles, acts as positive labels for the next step violation modeling. According to the FDR procedure, the positivity hypothesis is rejected if any of the bins is still significant. 

It is important to note the difference between \textit{fundamental positivity} and \textit{practical positivity}. The first violation will not be solved by more data collection with the same data distribution as it due to some underlying rule. The latter relates to a very small amount of samples, which can still cause problems even if theoretically not as infeasible, also known as "soft violations" \cite{karavani2019discriminative}. Soft violations effectively increase the number of positivity violations artificially. By contrast, we aim to detect fundamental positivity violation via a statistical test, as described above, which reduces the number of violations rather than increases it. Thus, we leave it to the following causal inference method to estimate the confidence interval increase in feature space where soft violations occur. Our method can be adapted, if needed, to be soft by introducing a threshold below which histogram counts are regarded as noise. A statistical test in an environment in which there can be misclassifications or misdiagnoses, giving the possibility of nonzero counts from false positives even though no real examples may exist has been proposed \cite{bradley2013testing}. In our case these misclassification could be due to the propensity model errors and can be applied straightforwardly.

\subsection{Explainability}
\label{subsec:explanation}

\subsubsection{Violation Modeling}
\label{subsec:model}

Practically positivity violations occur due to business rules or regulations. For example, some segment of users might be non illegible for email marketing, or some patience must not get the treatment. These rules are often manifests as a logic rules over inequality conditions. Thus, the hypothesis space for positivity violations might be well explained by a decision tree \cite{safavian1991survey}. Therefore, we use decision trees in order to characterize the previously detected areas of violation. A decision tree is constructed to differentiate between areas of violation and non-violation as label over the same set (treatment or control). Constructing a decision tree only on the same set allows us to sidestep the differences between the treatment and control sets, which we know exist as we are not in an RCT setting. Two decision trees are therefore used - one for the control set and one for the treated set, with each one trained on its own positivity violations. It is importance to note the inherit trade-off between explainability and predictability. It is straight-forward to plot the model's performance in terms of predictability (e.g. AUC) vs. the generated positivity explainability complexity (e.g. number of inequalities, see next section \ref{subsec:prune}). In principle, this would allow to automatically adjust for the model complexity, e.g. tree-depth.

\subsubsection{Pruning}
\label{subsec:prune}

In order to make the underlying violation analysis interpretable by a human analyst and remove over-fitted rules, we need to simplify the generated tree from the previous step. There is only interest in explaining violations no interest in explaining areas of positivity and most of the data is not expected to be violating positivity, and as such the pruning method is asymmetrical between positivity and nonpositivity. The pruning works by making nodes that meet one of the following criteria into leaves:
\begin{itemize}
\item $\beta = 90\%$ of this node is non-positive.
\item Less than $\gamma = 1\%$ of the entire non-positive set is contained in this leaf.
\end{itemize}

The first criterion prevents the over-fitting of rule-sets which were already found to be sufficient, while the second one discards areas of non-positivity that are too small to be meaningful.  

\begin{algorithm}
\caption{Create a pruned decision tree for positivity explanation.}
\begin{algorithmic}
\Require $X$ - Dataset of features with the same treatment value, $V$ data points which violate positivity, $\beta$ parameter to prevent over-fitting, $\gamma$ parameter to discard irrelevant nodes
\Ensure $\mathcal{T}$ - A pruned decision tree
\State $\mathcal{T} \gets$ \texttt{Construct\_Tree$(X\setminus V,V)$} \Comment{Create a decision tree}
\State $n \leftarrow$ \texttt{root}$(\mathcal{T})$ 
\While{$n$ is not a leaf}
    \If{$\frac{n[V]}{n[X\setminus V]+n[V]}>\beta$} \Comment{$n[\cdot]$ - the amount of data points within a node}
        \State make $n$ leaf
    \EndIf
    \If{$n[V]<\gamma|V|$}
        \State make $n$ leaf
    \EndIf
    \If{$n$ \textbf{not} leaf}
        \State run loop with $n$ as left child
        \State run loop with $n$ as right child
    \EndIf
\EndWhile

\end{algorithmic}
\label{alg:find}
\end{algorithm}

\subsubsection{Explainable Text Generation}
\label{subsec:exp_text}
In order to explain the rules in a human-readable manner, we use a major property of decision trees - their explainability. The path to any leaf within a decision tree is a series of rules in the form of a specific variable being either lesser or greater than a value. These rules can be easily collected using the following algorithm:

\begin{algorithm}
\caption{Explain non-positivity in a given dataset}
\begin{algorithmic}
\Require $\mathcal{T}$ a pruned decision tree, a desired leaf $l$
\Ensure A set of rules $R$ defining $l$ 
\State Initialize an empty set $R$
\State Traverse from root node $T(root)$ to leaf $l$:
\For{each node in the path $n$}
    \State Let $X_i$ be the covariate inquired for split on node $k$
    \State Let $\vee \in \{\le, >\}$ be the sign used to split
    \State Let $c$ be the cutoff for that split
    \State Update $R \gets R \wedge (X_{i}\vee c)$ \Comment{$\vee$ is a comparison operator, $\le$ or $>$}
\EndFor
\end{algorithmic}
\label{alg:explain}
\end{algorithm}

The rules can then be translated into human-readable speech easily by following the format of "[covariate] is [lesser/greater] than [amount]".

\section{Results}
\begin{figure}[ht]
    \centering
    \includegraphics[width=\textwidth]{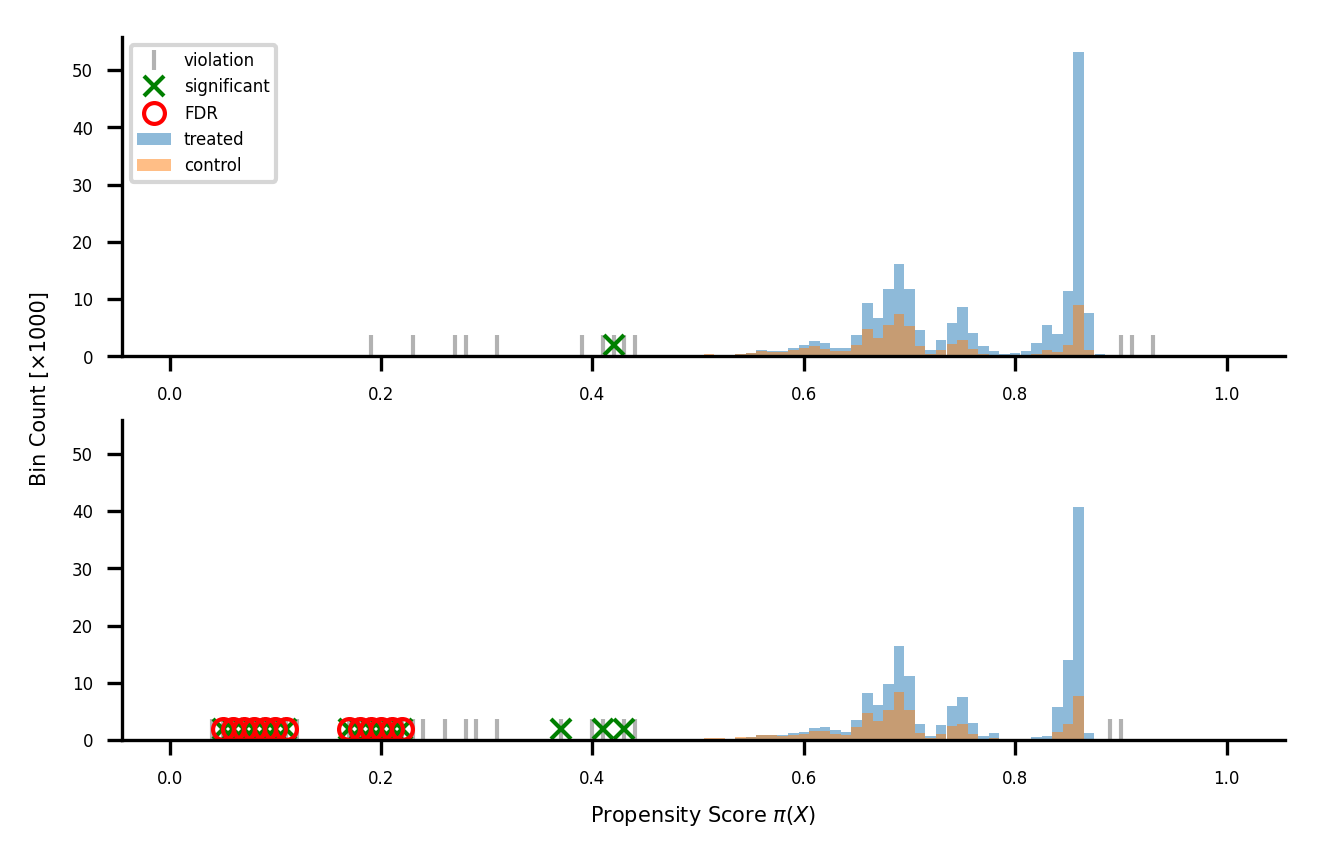}
    \caption{Positivity Violation Detection. The treatment is modeled via a propensity model. The resultant propensity score distribution is estimated using a histogram, for the control (orange) and treated (blue). Then, the bins in which violation occurs (probability is exclusively zero for one of the distribution) are detected (black ticks). The violated bins are then tested for statistical significance using $Z$-test proportion test to produce a list of $p$-values (significant bins, $\alpha=1\%$ are marked with a green $X$). Last the $p$-values are corrected for multiple testing using the false discovery rate (FDR) producing shorter significant bins (red circles). The latter is used as a positive label in the decision tree model of positivity violation used for explainability. In the top panel the analysis is conducted on real email campaign of a large software company. In the bottom panel the analysis is repeated after we synthetically removed a segment of people whose profile age was between 1500 and 1800 days old and experienced less than 50 days from the last email sent from the exposure group.}
    \label{fig:fdr}
\end{figure}

\begin{figure}[ht]
    \centering
    \includegraphics[scale=0.07]{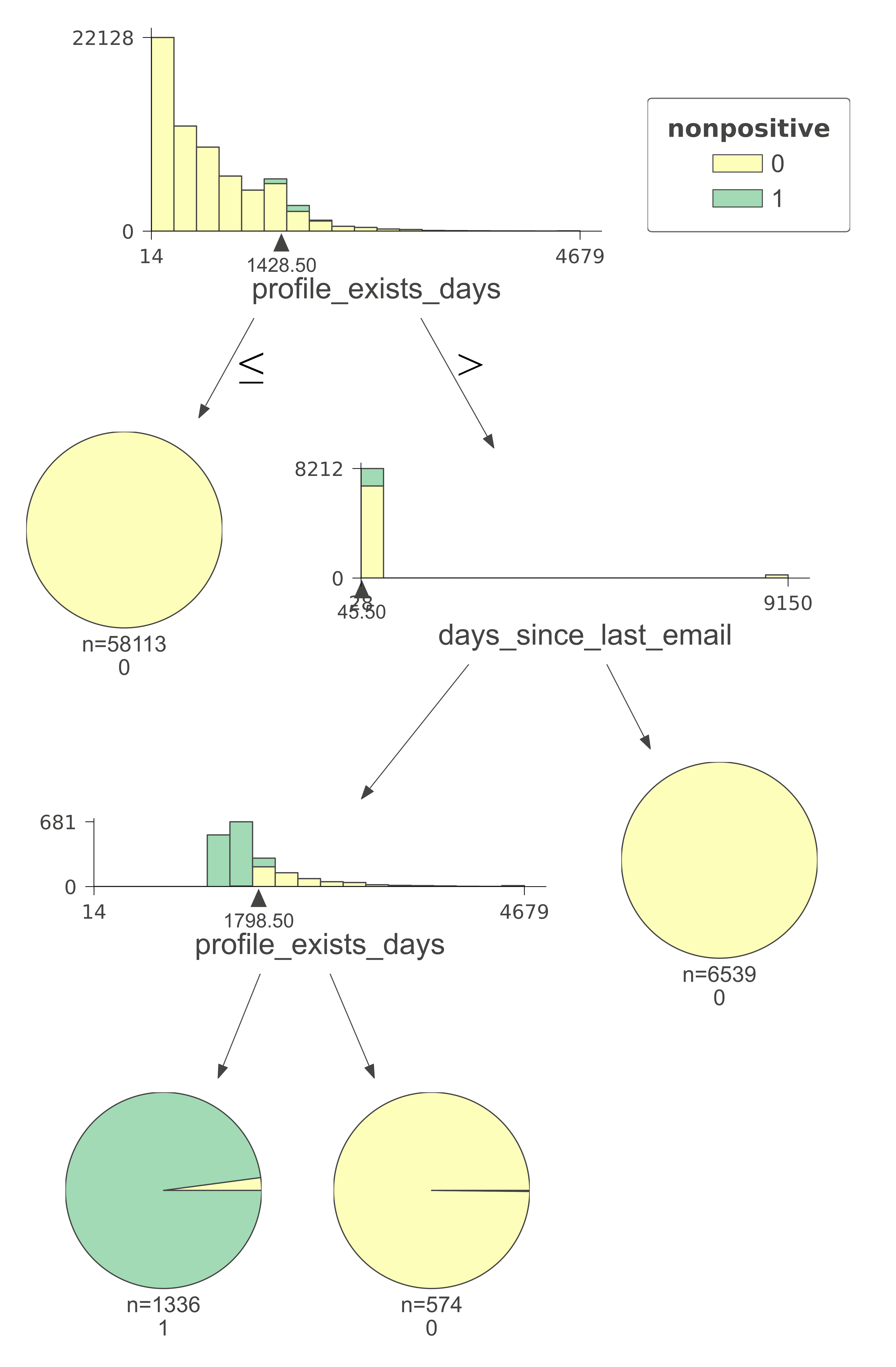}
    \caption{Illustration for the explainability of the positivity using a decision tree}
    \label{fig:exp_vis}
\end{figure}

\label{sec:results}
In this section, we applied the method from section \ref{sec:ourmethod}, to a real dataset of a large software enterprise company sending emails to its customers. We synthetically removed a segment of people whose profile age was between 1500 and 1800 days old and experienced less than 50 days from the last email sent from the exposure group. Figure \ref{fig:fdr} shows that the algorithm detects the regions in the data that have no positivity and recovers the underlying data generation process. Figure \ref{fig:exp_vis} shows a pruned decision tree representing those rules, which can be easily converted into text. Thus, a business analyst can either understand the observational analysis is not valid to the above segment or decide this is an area where further data collection is needed.

\section{Conclusion}
In this paper, we presented an algorithm to detect and explain positivity violation on a given dataset. The method models the propensity scores of the dataset, estimates the distribution of propensities and then uses statistical tests to detect positivity violations. A pruned decision tree is then constructed in order to provide an easily-understood explanation of the areas which lack positivity in the dataset. We demonstrate our method by applying it to a large dataset of a software enterprise company. Such methods of positivity detection and explanation are currently nonexistent within the causal inference open-source space, to the best of our knowledge. We hope a way to detect and explain areas of nonpositivity will prove useful to both researchers and laymen conducting causal inference experiments. Further work may be conducted in the generation of propensity scores and the estimation of the histogram through methods such as kernel smoothing which might take into account the propensity model posterior. Better metrics for the comparison of algorithms in the field will also be helpful to assess the strength of positivity-explanation methods. The inclusion of positivity-detection problems in future causal inference competitions \cite{dorie2019automated} will also help to create benchmarks for comparisons of different methods of positivity analysis.

\bibliographystyle{unsrt}
\bibliography{main}  
\end{document}